\documentclass{article} % For LaTeX2e
\usepackage{iclr2025_conference,times}
\usepackage{natbib}

\usepackage{amsmath,amsfonts,bm}

\def\eqref#1{equation~\ref{#1}}
\def\1{\bm{1}}

\DeclareMathAlphabet{\mathsfit}{\encodingdefault}{\sfdefault}{m}{sl}
\SetMathAlphabet{\mathsfit}{bold}{\encodingdefault}{\sfdefault}{bx}{n}

\usepackage{hyperref}
\usepackage{url}
\usepackage{algorithm}
\usepackage{algpseudocode}
\usepackage{graphicx}
\usepackage{booktabs}
\usepackage{amsmath}
\usepackage{amssymb}
\usepackage{cleveref}
\title{Preference Tree Optimization: Enhancing Goal-Oriented Dialogue with Look-Ahead Simulations}

\author{Lior Baruch\\
School of Computer Science\\
Reichman University, Herzliya, Israel\\
\texttt{lior95bar@gmail.com}
\And
Moshe Butman\\
Faculty of Computer Science\\
The College of Management Academic Studies, Rishon LeZion, Israel\\
\texttt{moshe.butman@gmail.com}
\And
Kfir Bar\\
School of Computer Science\\
Reichman University, Herzliya, Israel\\
\texttt{kfir.bar@runi.ac.il}
\And
Doron Friedman\\
School of Communications\\
Reichman University, Herzliya, Israel\\
\texttt{doronf@runi.ac.il}
}

\date{\today}

\iclrfinalcopy % Uncomment for camera-ready version, but NOT for submission.

\begin{document}

\maketitle
\begin{abstract}
Developing dialogue systems capable of engaging in multi-turn, goal-oriented conversations remains a significant challenge, especially in specialized domains with limited data. This research proposes a novel framework called \textit{Preference Tree Optimization (PTO)}, designed to iteratively improve agent models in such dialogue systems, by generating preference data using a method called \textit{Preference Tree with Look-Ahead}. Focusing on Motivational Interviewing (MI)—a counseling technique aimed at facilitating behavioral change—we leverage virtual patients and an oracle evaluator to simulate conversations and generate rich preference datasets. By combining this method with Direct Preference Optimization (DPO), we aim to enhance the agent's decision-making capabilities over iterative training cycles. The proposed framework addresses data scarcity and advances the development of more nuanced and effective dialogue systems in goal-oriented domains.

Experimental evaluations demonstrate that the PTO framework enhances dialogue agents' performance in goal-oriented conversations within the domain of Motivational Interviewing (MI). Models trained with PTO consistently outperformed the baseline in key metrics such as session satisfaction and working alliance. Additionally, incorporating look-ahead simulations led to improved long-term planning and more effective conversational strategies, with deeper look-ahead configurations yielding the most stable and high-scoring results.

\end{abstract}

\section{Introduction}

Goal-oriented dialogue systems are designed to achieve specific objectives through interactive conversations. Developing such systems in specialized domains is challenging due to the complexity of interactions and the scarcity of domain-specific data. Motivational Interviewing (MI) is such a domain -- it is a counseling approach that facilitates behavioral change through collaborative, client-centered dialogue, requiring nuanced understanding and adaptability from the conversational agent \cite{miller:1991:motivational}.

\begin{figure}[H]
  \centering
  \includegraphics[width=\linewidth]{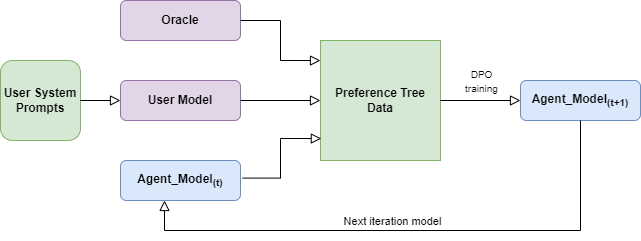}
  \caption{
   \textbf{Preference Tree Optimization (PTO) Framework}. The framework operates in two iterative steps: (i) \textbf{Preference Data Generation}: The User Model is prompted with a range of attributes to simulate diverse user personalities. For each digital user personality, the \textit{Preference Tree with Look-Ahead} (Section \ref{sec:PTO}) method is used in conjunction with the Oracle Evaluator and the current agent model (\(Agent\_Model_{t}\)) to generate a preference tree that explores various conversational pathways. These trees are aggregated into a comprehensive preference dataset. (ii) \textbf{Model Training}: The current agent model is trained on the newly generated preference dataset using Direct Preference Optimization (DPO), resulting in an improved model (\(Agent\_Model_{t+1}\)). The updated agent model is then used for the next iteration, repeating the process for continuous improvement.
  }
  \label{fig:framework}
\end{figure}

This research introduces a framework for iteratively improving agent models in goal-oriented dialogue systems, called \textit{Preference Tree Optimization (PTO)} (see~\Cref{fig:framework}), by generating preference data using a novel method called \textit{Preference Tree with Look-Ahead}. This method systematically simulates various conversational paths and evaluates them using an oracle to generate preference data. We use this preference data with Direct Preference Optimization (DPO)~\cite{rafailov:2023} to iteratively refine the agent model, enhancing its decision-making capabilities.

%Existing methods for dialogue system optimization often require intensive online computations. In contrast, our approach is designed for offline training: once the automated therapist is trained using the PTO framework, it can be deployed for real-time interactions without incurring the computational overhead of DPO at every conversation turn.

Our approach leverages existing virtual patients and evaluators from previous research in MI~\cite{yosef:2024}, making it an ideal testbed for our framework. By addressing the challenges of data scarcity and the need for nuanced interactions, we aim to contribute to the advancement of dialogue systems capable of effective, goal-oriented conversations.

Similar preference-based strategies have improved models in well-defined analytic tasks like games, coding, and math. However, their application to human-centric domains like Motivational Interviewing—where objectives are subjective and nuanced communication is key—remains largely unexplored.

%Figure 1 lacks clarity and does not fully capture the underlying mechanism of the proposed method. It should be improved. 

The PTO framework is designed exclusively as an offline training paradigm. Although the DPO process is computationally intensive—since it is applied at each simulated decision point during training—this cost is incurred only once during model development. Once trained, the automated therapist is deployed for real-time conversation, where the inference process is fast and efficient. 

\paragraph{}  
This work makes several contributions to the advancement of goal-oriented dialogue systems. First, we introduce the \textit{Preference Tree with Look-Ahead}, a novel method that systematically simulates and evaluates potential conversational trajectories to generate high-quality preference data, thereby facilitating more effective learning from interactions. Second, we propose the \textit{Preference Tree Optimization (PTO)} framework, which integrates this preference data with \textit{Direct Preference Optimization (DPO)} to iteratively refine an agent’s decision-making capabilities over successive training cycles. Third, we validate our approach in the challenging domain of \textit{Motivational Interviewing (MI)} by leveraging virtual patients and oracle evaluators to simulate realistic, high-stakes conversational scenarios. Finally, our methodology offers broad insights and generalizable strategies for applying preference-based optimization to other specialized dialogue domains.

\section{Background and Related Work}

Recent breakthroughs in Natural Language Processing (NLP) and the development of Large Language Models (LLMs) have dramatically advanced dialogue systems. However, designing goal-oriented systems for specialized domains—such as Motivational Interviewing (MI)—remains particularly challenging due to the limited availability of domain-specific data and the complexity of managing nuanced, multi-turn interactions. Pure generative models, which primarily rely on likelihood estimation, may not naturally exhibit goal-directed behavior. While reinforcement learning (RL) offers a potential path to integrating goal orientation, identifying suitable reward functions in domains like psychology is far from straightforward—unlike more structured fields such as mathematics or gaming, where clear optimal strategies exist. Additionally, approaches like Direct Preference Optimization (DPO) raise questions: Can they sufficiently promote goal-oriented behavior, and if so, what implicit reward mechanisms do they employ?

\subsection{Preference Optimization in Language Models}

One of the key approaches to improving language models involves aligning them with human preferences. This alignment helps models generate responses that are not only coherent, but also contextually appropriate and tailored to specific conversational objectives. Traditional approaches, such as Reinforcement Learning from Human Feedback (RLHF)~\cite{christiano:2017}, involve training a separate reward model based on human evaluations of model outputs. This reward model then guides the language model through reinforcement learning to produce preferred responses. Although effective, RLHF can be complex and resource-intensive due to the necessity of maintaining a distinct reward model and implementing reinforcement learning algorithms~\cite{ouyang:2022}.

Direct Preference Optimization (DPO)~\cite{rafailov:2023} offers a more streamlined alternative by directly optimizing the language model using preference data, eliminating the need for a separate reward model and the complexities of reinforcement learning. DPO establishes a direct mapping between LLM policies and reward functions, enabling the training of an LLM to satisfy preference data through a straightforward cross-entropy loss.

\subsection{Synthetic Data Generation and Iterative Self-Improvement}

Addressing the challenge of data scarcity in specialized dialogue domains has motivated researchers to develop methods that combine synthetic data generation with iterative self-improvement. Broadly, these approaches can be grouped into three categories: score-based synthetic data generation, self-evaluation–driven improvement, and search-based tree-structured methods.

\textbf{Score-Based Synthetic Data Generation:}  
Pace et al.~\cite{pace:2024} introduced \textit{West-of-N}, a method that leverages language models to produce multiple candidate responses for a given prompt. A reward model then scores these responses, and by selecting the best and worst outputs, the approach forms synthetic preference pairs used to refine the reward model’s alignment with human preferences. In a similar vein, ~\cite{guo:2024:OnlineDPO} propose an online variant of direct alignment from preferences. Their method employs an LLM as an annotator to provide on-the-fly feedback on pairs of responses sampled from the current model. This Online AI Feedback (OAIF) approach addresses the distribution shift inherent in offline datasets by continuously updating preference data, thereby enhancing alignment performance, particularly in soft domains where nuanced judgment is critical.

\textbf{Self-Evaluation–Driven Improvement:}  
Another line of work harnesses the model’s internal evaluation mechanisms to self-generate and refine synthetic data. Yuan et al.~\cite{yuan:2024:SelfReward} present \textit{Self-Rewarding Language Models}, wherein the model generates multiple responses and uses an LLM-as-a-judge to rank them. The resulting preference data is then used with Direct Preference Optimization (DPO) to iteratively enhance both response generation and internal reward estimation. Similarly, Liang et al.~\cite{liang:2024} propose \textit{I-SHEEP} (Iterative Self-EnHancEmEnt Paradigm), a framework in which the model synthesizes data, self-assesses its quality, and filters out low-quality responses before applying supervised fine-tuning. While these self-assessment–based methods efficiently leverage the model’s own capabilities, they risk perpetuating internal biases if the self-evaluation is not sufficiently robust.

\textbf{Search-Based and Tree-Structured Approaches:}
A third category of methods employs search strategies to systematically explore potential outcomes. Xie et al.~\cite{xie:2024} integrate Monte Carlo Tree Search (MCTS) with iterative preference learning to generate and evaluate fine-grained, step-level reasoning paths. The collected preference data is then used to refine the model via DPO. In addition, prior work on \textit{Preference Trees}~\cite{yuan:2024:PrefTree} demonstrates how tree-structured methods can effectively manage complex reasoning tasks in domains such as coding, math, and logic. Yu et al.~\cite{yu2023prompt} introduce a prompt-based search method where an LLM plays multiple roles in planning without additional training. Recently, Chen et al.~\cite{chenbroaden} developed a conversation planning approach that reduces reliance on direct LLM-based simulation, by exploiting the dense semantic representation of conversations. Building on these ideas, our framework employs a \textit{Preference Tree with Look-Ahead} to simulate full conversational trajectories using a dedicated user model, specifically targeting goal-oriented dialogue systems like those used in Motivational Interviewing.  

\noindent In summary, these diverse methodologies illustrate the potential of combining synthetic data generation with iterative self-improvement. They differ in how preference data is generated, whether through score-based selection, self-assessment, or search-based exploration, and in the application domains they target. Our work bridges search-based and score-based paradigms: the \textit{Preference Tree with Look-Ahead} method employs tree-structured exploration of conversational trajectories, while the oracle evaluator provides score-driven comparisons, enabling iterative refinement via \textit{Direct Preference Optimization (DPO)} to enhance goal-oriented dialogue agents in specialized domains such as \textit{Motivational Interviewing}.
Unlike prior work, which has primarily focused on structured tasks such as coding, math, or games, our approach explores preference-based optimization in a domain that requires deep human understanding, where objectives are inherently subjective and harder to quantify.

\subsection{Motivational Interviewing and AI Dialogue Systems}

Motivational Interviewing (MI) is a client-centered counseling approach aimed at eliciting behavioral change by helping clients explore and resolve ambivalence~\cite{miller:1991:motivational}. Implementing MI in AI dialogue systems presents unique challenges due to the need for empathy, adaptability, and the ability to interpret subtle conversational cues.

Previous research has explored the potential of LLMs in simulating MI sessions. Yosef et al.~\cite{yosef:2024} utilized AI-generated patient simulations to assess MI sessions, highlighting the feasibility of virtual patients in training and evaluating therapeutic dialogues. Their work demonstrated that AI agents could engage in MI conversations to a certain extent but also underscored the limitations in capturing the full depth of human therapist-patient interactions.

In addition, Yosef et al. fine-tuned therapist models using existing datasets specific to MI, demonstrating that such fine-tuning can improve model performance in therapeutic settings~\cite{yosef:2024}. Unlike methods that rely on pre-existing datasets, our \textit{Preference Tree Optimization (PTO)} framework iteratively generates training data from simulated conversations using the \textit{Preference Tree with Look-Ahead} method, refining the model at each iteration via Direct Preference Optimization.

\section{Method}

Our methodology involves two main components: the \textit{Preference Tree with Look-Ahead} method for preference data generation and an iterative training process to refine the agent model using DPO.

\subsection{Preference Tree with Look-Ahead}
\label{sec:PTO}

The \textit{Preference Tree with Look-Ahead} method systematically explores potential conversational paths by simulating multiple agent responses and their subsequent dialogue trajectories, as shown in \Cref{sec:alg_prefTree}. This is intended to allow the agent to anticipate the long-term impact of its responses. The process is as follows:

\begin{enumerate}
    \item \textbf{Agent Decision Point}: At each turn, the agent model generates $N$ possible responses.
    \item \textbf{Branch Initialization}: For each response, a new branch is created, and the response is appended to the conversation history.
    \item \textbf{Look-Ahead Simulation}: Each branch simulates $K$ future steps, alternating between the agent and the virtual patient, to anticipate the long-term implications of the agent's response.
    \item \textbf{Oracle Evaluation}: An oracle evaluator assesses each branch based on predefined criteria (e.g., adherence to MI principles, empathy, goal progression) and assigns scores.
    \item \textbf{Preference Recording}: The response with the highest score is considered the preferred response, and the one with the lowest score is the least preferred. The preference tuple is recorded in the dataset.
    \item \textbf{Conversation Update}: The conversation continues with the preferred response, and the process repeats until a termination condition is met (e.g., reaching maximum conversation length or achieving the goal).
\end{enumerate}

\begin{figure}[H]
  \centering
  \includegraphics[width=0.565\linewidth]{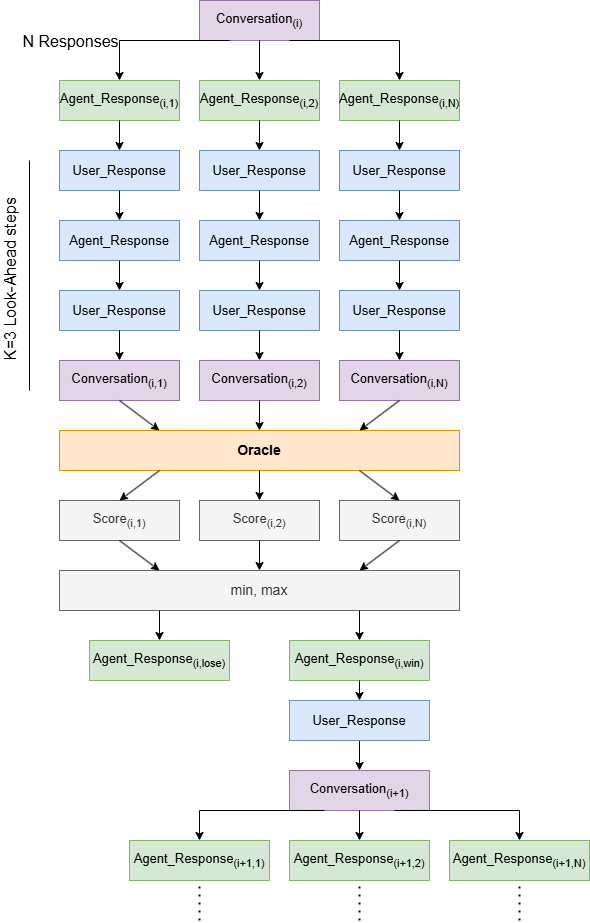}
  \caption{
   \textbf{Preference Tree Generation Process.} The figure shows how a preference tree is used to generate preference data. At each conversation step \(i\), the agent generates \(N\) possible responses, and each branch simulates the conversation through several look-ahead steps. These branches represent possible future dialogue paths. An oracle evaluates each path, assigning scores to determine the best (\(response_{i,win}\)) and worst (\(response_{i,lose}\)) outcomes. After selecting the winning response, the user model replies, advancing to the next conversation step \(conversation_{i+1}\), and the process repeats. This way, each preference tree produces multiple preference samples, with each sample consisting of a tuple \((conversation_i, response_{i,lose}, response_{i,win})\).
  }
  \label{fig:Preference Tree}
\end{figure}

By considering future conversation trajectories, the agent is expected to learn to make decisions that are not only immediately appropriate but also beneficial in the long term (see~\Cref{fig:Preference Tree}).

\subsection{Preference Tree Optimization (PTO) Framework}

This process forms the Preference Tree Optimization (PTO) Framework.
The agent model is iteratively improved through cycles of preference data generation and training using DPO.

\begin{enumerate}
    \item \textbf{Initial Training}: The agent model is initially trained on available data or pre-trained weights.
    \item \textbf{Preference Data Generation}: Using the current agent model, the \textit{Preference Tree with Look-Ahead} method generates new preference data, capturing the agent's strengths and weaknesses.
    \item \textbf{Preference Data Filtering}: We retain a preference sample only if the winning score surpasses the losing score by a predefined threshold. In our experiments, we used a threshold value of 0.1, ensuring that only clearly distinguishable preference pairs contribute to training.
    \item \textbf{Model Update}: The agent model is fine-tuned using DPO on the newly generated preference data, optimizing it directly based on preferences without the need for a reward model.
    \item \textbf{Evaluation}: The updated model is evaluated using predefined metrics to assess improvements.
    \item \textbf{Iteration}: Steps 2-5 are repeated, allowing the agent to improve over time through continuous learning.
\end{enumerate}

This process balances exploration (generating new conversational paths) and exploitation (refining the agent's responses), leading to incremental enhancements in performance.

\begin{algorithm}[H]
\caption{Preference Tree Optimization (PTO) Framework}
\label{alg:PTOF}
\begin{algorithmic}[1]
\Require Initial agent model $A^{(0)}$, user model $U$, oracle evaluator $O$, maximum conversation length $L$, look-ahead steps $K$, branching factor $N$, trees per iteration $T$, total iterations $I$, filtering threshold $\tau$
\Ensure Sequence of optimized agent models $\{A^{(1)}, A^{(2)}, \dots, A^{(I)}\}$
\For{$i \gets 1$ \textbf{to} $I$}
    \State Initialize preference dataset: $D^{(i)} \gets \varnothing$
    \For{$t \gets 1$ \textbf{to} $T$}
        \State Assign user role: $U_t \gets U$
        \State $P^{(t)} \gets \textbf{GeneratePreferenceTree}(A^{(i-1)}, U_t, O, L, K, N)$ \Comment{See \Cref{alg:preference_tree}}
        \State Aggregate preferences: $D^{(i)} \gets D^{(i)} \cup P^{(t)}$
    \EndFor
    \State $D^{(i)} \gets \textbf{Filter}(D^{(i)}, \tau)$ \Comment{Retain samples where the winning score exceeds the losing score by at least $\tau$}
    \State $A^{(i)} \gets \textbf{DPO}(A^{(i-1)}, D^{(i)})$
\EndFor
\State \Return Optimized agent models $\{A^{(1)}, A^{(2)}, \dots, A^{(I)}\}$
\end{algorithmic}
\end{algorithm}

\section{Experimental Setup}

During our experiments, GPT-3.5 served as both the user simulator and the oracle evaluator. Notably, GPT-3.5 was used in fixed, separate roles (with distinct prompts for the user simulation and the oracle evaluation), and it was not updated or fine-tuned at any point during the training process. This ensured that the model’s parameters remained unchanged throughout, providing a consistent but unlearned behavior in each role.

To evaluate our proposed framework, we conducted a series of initial experiments in the Motivational Interviewing (MI) domain. The experimental setup is detailed as follows:

\subsection{Models and Tools}

\begin{itemize}
    \item \textbf{Agent Model}: We utilized \textit{Llama-2-7B} as the base model for the therapist agent.
    \item \textbf{User Model}: Virtual patients were simulated using \textit{GPT-3.5}, based on guidelines from previous MI research \cite{yosef:2024}. Each patient is defined by parameters such as gender, age, problem (smoking/obesity), duration, prior attempts to resolve the issue, and cooperation level, creating 96 unique profiles to capture diverse challenges and attitudes toward counseling. 
    \item \textbf{Oracle Evaluator}: GPT-3.5 model is used as the oracle evaluator, using specific questionnaires designed to assess MI adherence and conversational quality based on the guidelines from previous research~\cite{yosef:2024} and detailed in \Cref{sec:Questionnaire}. The final score is calculated as the average of the two questionnaire scores, where each questionnaire score is the average of its respective question scores.

\end{itemize}

\subsection{Experimental Variables}

\begin{itemize}
    \item \textbf{Look-Ahead Depths}: We tested two different look-ahead depths: 0 (no look-ahead) and 5. This variable assesses the impact of anticipating future conversational turns on the agent's performance.
    \item \textbf{Iterations per Look-Ahead}: For each look-ahead depth, we conducted 7 iterative training cycles. Each iteration involved:
    \begin{enumerate}
        \item \textbf{Preference Data Generation}: Utilizing the \textit{Preference Tree with Look-Ahead} method to generate preference tuples based on simulated conversational paths.
        \item \textbf{Model Fine-Tuning}: Applying Direct Preference Optimization (DPO) to fine-tune the agent model using the newly generated preference data.
    \end{enumerate}
\end{itemize}

\subsection{Data Collection}

After each iteration, we generated a set of conversations to evaluate the agent's performance:
\begin{itemize}
    \item \textbf{Number of Conversations}: For each trained model, we conducted 96 separate conversations with virtual patients to ensure a comprehensive assessment.
    \item \textbf{Evaluation Metrics}: Each conversation was scored by the oracle evaluator based on two distinct questionnaires designed to measure MI adherence and overall conversational quality, detailed in Table~\ref{Questionnaires}.
\end{itemize}

\section{Results}

To assess the efficacy of the proposed Preference Tree Optimization (PTO) Framework, we conducted experiments concentrating on two distinct look-ahead depths: 0 and 5. Each configuration was subjected to seven iterative training cycles, and their performances were compared against the baseline model, \textit{Llama-2-7B}.

\subsection{Performance Metrics}

The agent's effectiveness was evaluated using two primary metrics derived from the oracle evaluator's questionnaires (see~\Cref{Questionnaires}):

\begin{itemize}
    \item \textbf{Session Satisfaction (Q1)}: This metric aggregates scores from Questionnaire 1 (as detailed in~\cite{yosef:2024}), assessing overall satisfaction, content relevance, motivation facilitation, learning outcomes, and applicability to everyday life.
    \item \textbf{Working Alliance (Q2)}: This metric aggregates scores from Questionnaire 2 (see~\cite{yosef:2024}), evaluating the therapist's interpersonal skills, empathy, communication effectiveness, and ability to establish a collaborative relationship.
    \item \textbf{Final Score}: Calculated as the average of Session Satisfaction and Working Alliance scores, this provides a comprehensive indicator of overall performance.
\end{itemize}

\subsection{Results Overview}

\begin{table}[H]
    \centering
    \caption{Average Performance Scores and Standard Deviations Across Models}
    \begin{tabular}{lcccccc}
    
        \toprule
        \textbf{Model} & \multicolumn{2}{c}{\textbf{Session Satisfaction (Q1)}} & \multicolumn{2}{c}{\textbf{Working Alliance (Q2)}} & \multicolumn{2}{c}{\textbf{Final Score}} \\
        \cmidrule(lr){2-3} \cmidrule(lr){4-5} \cmidrule(lr){6-7}
         & Mean & SD & Mean & SD & Mean & SD \\
        \midrule
        \textit{Base} & 3.521 & 1.056 & 3.385 & 0.539 & 3.453 & 0.740 \\
        \midrule
        \multicolumn{7}{l}{\textbf{Look-Ahead Depth 0}} \\
        L0\_M1 & 3.863 & 1.012 & 3.452 & 0.731 & 3.657 & 0.824 \\
        L0\_M2 & 3.750 & 1.059 & 3.435 & 0.788 & 3.593 & 0.878 \\
        L0\_M3 & 3.796 & 0.868 & 3.567 & 0.511 & 3.682 & 0.649 \\
        L0\_M4 & 3.969 & 0.979 & 3.585 & 0.642 & 3.777 & 0.769 \\
        L0\_M5 & 3.744 & 1.124 & 3.478 & 0.687 & 3.611 & 0.856 \\
        L0\_M6 & 3.794 & 1.143 & 3.494 & 0.633 & 3.644 & 0.834 \\
        L0\_M7& 3.677& 1.098& 3.452& 0.667& 3.565&0.828\\
        \midrule
        \multicolumn{7}{l}{\textbf{Look-Ahead Depth 5}} \\
        L5\_M1 & 3.898 & 1.005 & 3.523 & 0.480 & 3.710 & 0.712 \\
        L5\_M2 & 3.969 & 0.809 & 3.618 & 0.455 & 3.794 & 0.594 \\
        L5\_M3 & 4.050 & 0.818 & 3.683& 0.548 & 3.866 & 0.611 \\
        L5\_M4 & 3.981 & 0.801 & 3.605 & 0.351 & 3.793 & 0.524 \\
        L5\_M5 & \textbf{4.225} & 0.775 & 3.660 & 0.451 & 3.942& 0.559 \\
        L5\_M6 & 4.112 & 0.868 & 3.656 & 0.477 & 3.884 & 0.629 \\
        L5\_M7& 4.190& \underline{0.614}& \textbf{3.775}& \underline{0.332}& \textbf{3.982}& \underline{0.414}\\
        \bottomrule
    \end{tabular}
    \label{tab:preliminary_results}
\end{table}

\Cref{tab:preliminary_results} presents the mean scores and standard deviations for \textbf{Session Satisfaction (Q1)}, \textbf{Working Alliance (Q2)}, and the \textbf{Final Score} across all evaluated models, including the baseline (\textit{Llama-2-7B}) and PTO-enhanced models at look-ahead depths of 0 and 5. The lowest standard deviation values for each metric are underlined in the table. 

\begin{figure}[H]
    \centering
    \includegraphics[width=1\linewidth]{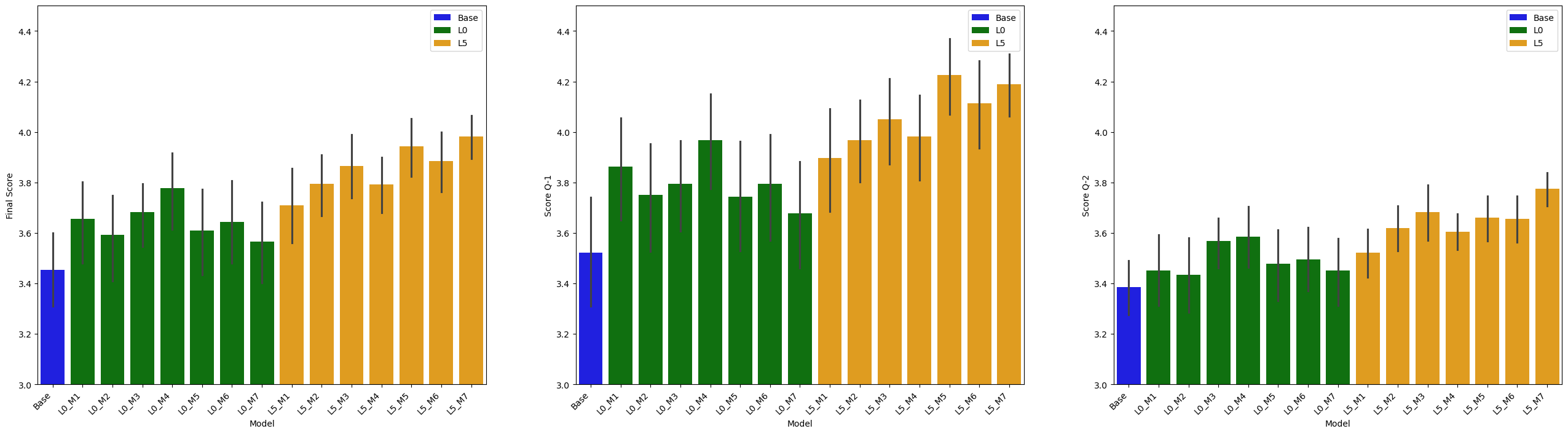}
    \caption{
        \textbf{Comparative Performance Analysis} \\
        Bar charts illustrating the average scores for the \textbf{Final Score} (left), \textbf{Session Satisfaction (Q1)} (middle), and \textbf{Working Alliance (Q2)} (right) across the \textit{Baseline} model (\textit{Llama-2-7B}) and the \textit{PTO-enhanced} models with varying look-ahead depths.
        Error bars represent the 95\% confidence intervals.
        This comparison highlights the performance improvements achieved through the Preference Tree Optimization Framework.
    }
    \label{fig:barplot_scores}
\end{figure}

Across all evaluated metrics, every PTO-trained model (L0\_Mx and L5\_Mx) outperforms the baseline (see~\Cref{fig:barplot_scores}), demonstrating that preference-based optimization improves goal-oriented dialogue performance. Additionally, models trained with deeper look-ahead (depth-5) achieve higher scores than those trained with no look-ahead (depth-0), suggesting that anticipating future conversational paths enhances both session satisfaction and the working alliance.
A one-way ANOVA confirms that model choice significantly influences Q1, Q2, Final Score, and conversation length as shown in \Cref{tab:anova}. Post-hoc Tukey HSD tests were conducted to compare the baseline model (\textit{Llama-2-7B}) against the best-performing models from each look-ahead depth: \textbf{L0\_M4} (best-performing depth-0 model) and \textbf{L5\_M7} (best-performing depth-5 model) (\Cref{sec:tukey}). Results indicate that both \textbf{L0\_M4} and \textbf{L5\_M7} significantly outperform the baseline across all three metrics (Q1, Q2, and Final Score). While L5\_M7 achieves the highest Final Score, its improvement over L0\_M4 is only statistically significant for Q2, indicating that deeper look-ahead particularly strengthens the working alliance.

Examining the standard deviations in Table~\ref{tab:preliminary_results} further supports the stability of PTO-trained models. Among all evaluated models, \textbf{L5\_M7} exhibits the lowest variance across Q1, Q2, and Final Score (underlined in the table), suggesting that deeper look-ahead not only enhances performance but also ensures more consistent and reliable motivation interventions.

\begin{figure}[H]
    \centering
    \includegraphics[width=0.7\linewidth]{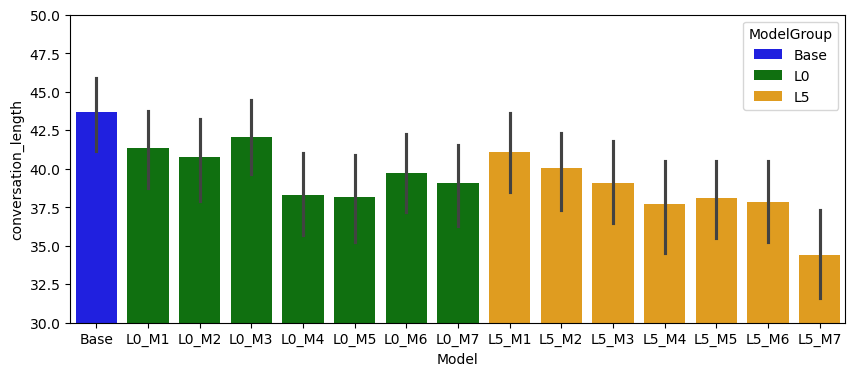}
    \caption{
        \textbf{Barplot of Conversation Length} \\
        This barplot displays the average conversation lengths for each model, comparing the \textit{Baseline} model with the \textit{PTO-enhanced} models at different look-ahead depths.
        Error bars represent the 95\% confidence intervals.
        It highlights how the Preference Tree Optimization Framework influences the efficiency and duration of dialogues.
    }
    \label{fig:barplot_conversation_length}
\end{figure}

Furthermore, Figure~\ref{fig:barplot_conversation_length} illustrates that PTO-trained models tend to reduce conversation length compared to the baseline, reflecting more focused interactions. Notably, \textbf{L5\_M7} achieves the most substantial reduction, decreasing the average number of dialogue turns from 43.7 (baseline) to 34.4. This underscores the role of look-ahead in streamlining interactions while maintaining high conversation quality.

\begin{table}[H]
    \centering
    \caption{One-Way ANOVA Results for Model Performance}
    \begin{tabular}{lcc}
        \toprule
        \textbf{Metric} & \textbf{F-Statistic} & \textbf{p-value} \\
        \midrule
        Final Score & 15.637 & 3.60e-07 \\
        Session Satisfaction (Q1) & 13.654 & 2.17e-06 \\
        Working Alliance (Q2) & 13.446 & 2.63e-06 \\
        Conversation Length & 11.928 & 1.06e-05 \\
        \bottomrule
    \end{tabular}
    \label{tab:anova}
\end{table}

\section{Discussion}

Our experimental results demonstrate that the Preference Tree Optimization (PTO) framework consistently improves dialogue performance compared to the baseline. Importantly, these improvements were achieved using a base pre-trained model (Llama-2-7B) that was neither instruction-tuned nor fine-tuned via supervised learning; instead, all training was conducted solely with data generated by the Preference Tree with Look-Ahead method. Both look-ahead configurations (depth-0 and depth-5) yield significant gains in Session Satisfaction (Q1), Working Alliance (Q2), and overall Final Score. Notably, the best-performing depth-5 model (L5\_M7) not only achieved the highest scores but also exhibited the lowest variance, indicating more stable and reliable interactions. This suggests that incorporating look-ahead enables the agent to anticipate future conversational turns, leading to more effective, empathetic, and streamlined dialogues.

Potential biases in automated evaluation remain a concern. For instance, positional bias may occur if the evaluator assigns different weights to responses depending on their position in the conversation. Our analysis shows that while there is minor variability in the evaluation of the initial utterances, the oracle’s scoring remains largely consistent throughout the dialogue. Similarly, preference bias can emerge if the evaluator favors certain stylistic or content-related features—such as preferring responses typical of language models over those created by humans—which could lead the agent to optimize for superficial attributes rather than genuine conversational quality. This can be considered a type of ``reward hacking''. 

In our case, both the oracle evaluator and the virtual patients are implemented as fixed, pre-trained models. Thus, the fact that we employ the same underlying model for both roles is not the primary source of risk for ``reward hacking''. Reward hacking is an inherent challenge in frameworks that rely on automated evaluation, regardless of whether identical or heterogeneous models are used. Importantly, our oracle evaluator was validated by human assessments—although the correlation was moderate, this validation indicates that the evaluation criteria capture meaningful aspects of effective counseling.

Future work will focus on elucidating whether and why deeper look-ahead (L5) offers advantages over no look-ahead (L0) in this soft domain. We plan to investigate the underlying mechanisms that contribute to improved long-term planning, such as better management of conversational dynamics and enhanced anticipatory decision-making. Additionally, we plan to benchmark our approach against leading state-of-the-art methods—specifically, the online alignment framework from~\cite{guo:2024:OnlineDPO} and the self-rewarding language model approach from~\cite{yuan:2024:SelfReward}—to further elucidate the advantages and limitations of long-term planning in goal-oriented dialogue.

 % \section{Research Road-map}

% \begin{enumerate}
%     \item \textbf{Phase 1 (May - June 2024))}: Develop the \textit{Preference Tree with Look-Ahead} algorithm and validate its ability to generate meaningful preference data.
%     \item \textbf{Phase 2 (July - August 2024)}: Implement the iterative training process using DPO and conduct initial experiments to assess improvements.
%     \item \textbf{Phase 3 (September - November 2024)}: Perform extensive evaluations against baselines using the defined metrics, analyzing the agent’s performance in MI dialogues.
%     \item \textbf{Phase 4 (December 2024 - February 2025)}: Refine the models and methods based on findings, explore scalability, and prepare for potential real-world applications.
%     \item \textbf{Phase 5 (March - April 2025)}: Publish results, share methodologies, and consider applications to other domains.
% \end{enumerate}

\section*{Acknowledgments}
This work was partially supported by the European Commission Horizon 2020 project GuestXR (\#101017884).

\newpage
\bibliography{iclr2025_conference}

\begin{thebibliography}{13}
\providecommand{\natexlab}[1]{#1}
\providecommand{\url}[1]{\texttt{#1}}
\expandafter\ifx\csname urlstyle\endcsname\relax
  \providecommand{\doi}[1]{doi: #1}\else
  \providecommand{\doi}{doi: \begingroup \urlstyle{rm}\Url}\fi

\bibitem[Chen et~al.(2025)Chen, Niu, Foo, and Low]{chenbroaden}
Zhiliang Chen, Xinyuan Niu, Chuan-Sheng Foo, and Bryan Kian~Hsiang Low.
\newblock Broaden your scope! efficient multi-turn conversation planning for
  llms with semantic space.
\newblock In \emph{The Thirteenth International Conference on Learning
  Representations}, 2025.

\bibitem[Christiano et~al.(2017)Christiano, Leike, Brown, Martic, Legg, and
  Amodei]{christiano:2017}
Paul~F. Christiano, Jan Leike, Tom~B. Brown, Miljan Martic, Shane Legg, and
  Dario Amodei.
\newblock Deep reinforcement learning from human preferences.
\newblock \emph{arXiv preprint arXiv:1706.03741}, 2017.
\newblock URL \url{https://arxiv.org/abs/1706.03741}.
\newblock Presented at the 31st Conference on Neural Information Processing
  Systems (NeurIPS 2017).

\bibitem[Guo et~al.(2024)Guo, Zhang, Liu, Liu, Khalman, Llinares, Ramé,
  Mesnard, Zhao, Piot, Ferret, and Blondel]{guo:2024:OnlineDPO}
Shangmin Guo, Biao Zhang, Tianlin Liu, Tianqi Liu, Misha Khalman, Felipe
  Llinares, Alexandre Ramé, Thomas Mesnard, Yao Zhao, Bilal Piot, Johan
  Ferret, and Mathieu Blondel.
\newblock Direct language model alignment from online ai feedback.
\newblock \emph{arXiv preprint arXiv:2402.04792}, 2024.
\newblock URL \url{https://arxiv.org/abs/2402.04792}.

\bibitem[Liang et~al.(2024)Liang, Zhang, Qu, Zheng, Guo, Du, Yang, Liu, Lin,
  Ma, Huang, and Zhang]{liang:2024}
Yiming Liang, Ge~Zhang, Xingwei Qu, Tianyu Zheng, Jiawei Guo, Xinrun Du,
  ZhenZhu Yang, Jiaheng Liu, Chenghua Lin, Lei Ma, Wenhao Huang, and Jiajun
  Zhang.
\newblock I-sheep: Self-alignment of llm from scratch through an iterative
  self-enhancement paradigm.
\newblock \emph{Proceedings of the AAAI Conference on Artificial Intelligence},
  2024.
\newblock URL \url{https://www.arxiv.org/abs/2408.08072}.
\newblock Copyright © 2024, Association for the Advancement of Artificial
  Intelligence (www.aaai.org). All rights reserved.

\bibitem[Miller \& Rollnick(1991)Miller and Rollnick]{miller:1991:motivational}
W.R. Miller and S.~Rollnick.
\newblock \emph{Motivational Interviewing: Preparing People to Change Addictive
  Behavior}.
\newblock Guilford Publications, 1991.
\newblock ISBN 9780898625660.
\newblock URL \url{https://books.google.co.il/books?id=h16_QgAACAAJ}.

\bibitem[Ouyang et~al.(2022)Ouyang, Wu, Jiang, Almeida, Wainwright, Mishkin,
  Zhang, Agarwal, Slama, Ray, Schulman, Hilton, Kelton, Miller, Simens, Askell,
  Welinder, Christiano, Leike, and Lowe]{ouyang:2022}
Long Ouyang, Jeff Wu, Xu~Jiang, Diogo Almeida, Carroll~L. Wainwright, Pamela
  Mishkin, Chong Zhang, Sandhini Agarwal, Katarina Slama, Alex Ray, John
  Schulman, Jacob Hilton, Fraser Kelton, Luke Miller, Maddie Simens, Amanda
  Askell, Peter Welinder, Paul Christiano, Jan Leike, and Ryan Lowe.
\newblock Training language models to follow instructions with human feedback.
\newblock \emph{arXiv preprint arXiv:2203.02155}, 2022.
\newblock URL \url{https://arxiv.org/abs/2203.02155}.
\newblock Work by the OpenAI team.

\bibitem[Pace et~al.(2024)Pace, Mallinson, Malmi, Krause, and
  Severyn]{pace:2024}
Aliz{\'e}e Pace, Jonathan Mallinson, Eric Malmi, Sebastian Krause, and Aliaksei
  Severyn.
\newblock West-of-n: Synthetic preference generation for improved reward
  modeling.
\newblock \emph{arXiv preprint arXiv:2401.12086}, 2024.
\newblock URL \url{https://arxiv.org/abs/2401.12086}.

\bibitem[Rafailov et~al.(2023)Rafailov, Sharma, Mitchell, Ermon, Manning, and
  Finn]{rafailov:2023}
Rafael Rafailov, Archit Sharma, Eric Mitchell, Stefano Ermon, Christopher~D.
  Manning, and Chelsea Finn.
\newblock Direct preference optimization: Your language model is secretly a
  reward model.
\newblock \emph{arXiv preprint arXiv:2305.18290}, 2023.
\newblock URL \url{https://arxiv.org/abs/2305.18290}.
\newblock Accepted at the 37th Conference on Neural Information Processing
  Systems (NeurIPS 2023).

\bibitem[Xie et~al.(2024)Xie, Goyal, Zheng, Kan, Lillicrap, Kawaguchi, and
  Shieh]{xie:2024}
Yuxi Xie, Anirudh Goyal, Wenyue Zheng, Min-Yen Kan, Timothy Lillicrap, Kenji
  Kawaguchi, and Michael Shieh.
\newblock Monte carlo tree search boosts reasoning via iterative preference
  learning.
\newblock \emph{arXiv preprint arXiv:2405.00451v2}, 2024.
\newblock URL \url{https://github.com/YuxiXie/MCTS-DPO}.

\bibitem[Yosef et~al.(2024)Yosef, Zisquit, Cohen, Klomek, Bar, and
  Friedman]{yosef:2024}
Stav Yosef, Moreah Zisquit, Ben Cohen, Anat~Brunstein Klomek, Kfir Bar, and
  Doron Friedman.
\newblock The journey towards an automatic mental health therapist.
\newblock \emph{Preprint}, 2024.

\bibitem[Yu et~al.(2023)Yu, Chen, and Yu]{yu2023prompt}
Xiao Yu, Maximillian Chen, and Zhou Yu.
\newblock Prompt-based monte-carlo tree search for goal-oriented dialogue
  policy planning.
\newblock \emph{arXiv preprint arXiv:2305.13660}, 2023.

\bibitem[Yuan et~al.(2024{\natexlab{a}})Yuan, Cui, Wang, Ding, Wang, Deng,
  Shan, Chen, Xie, Lin, Liu, Zhou, Peng, Liu, and Sun]{yuan:2024:PrefTree}
Lifan Yuan, Ganqu Cui, Hanbin Wang, Ning Ding, Xingyao Wang, Jia Deng, Boji
  Shan, Huimin Chen, Ruobing Xie, Yankai Lin, Zhenghao Liu, Bowen Zhou, Hao
  Peng, Zhiyuan Liu, and Maosong Sun.
\newblock Advancing llm reasoning generalists with preference trees.
\newblock \emph{arXiv preprint arXiv:2404.02078}, 2024{\natexlab{a}}.
\newblock URL \url{https://arxiv.org/abs/2404.02078}.
\newblock Preprint.

\bibitem[Yuan et~al.(2024{\natexlab{b}})Yuan, Pang, Cho, Sukhbaatar, Xu, and
  Weston]{yuan:2024:SelfReward}
Weizhe Yuan, Richard~Yuanzhe Pang, Kyunghyun Cho, Sainbayar Sukhbaatar, Jing
  Xu, and Jason Weston.
\newblock Self-rewarding language models.
\newblock \emph{arXiv preprint arXiv:2401.10020}, 2024{\natexlab{b}}.
\newblock URL \url{https://arxiv.org/abs/2401.10020}.

\end{thebibliography}
\bibliographystyle{iclr2025_conference}

\appendix
\section{Appendix}

\subsection{Preference Tree with Look-Ahead Algorithm}
\label{sec:alg_prefTree}
\begin{algorithm}[H]
\caption{Preference Tree with Look-Ahead}
\label{alg:preference_tree}
\begin{algorithmic}[1]
\small
\Require
    \begin{itemize}
        \item Agent model $A$
        \item User model $U$
        \item Oracle evaluator $O$
        \item Maximum conversation length $L$
        \item Look-ahead depth $K$
        \item Number of candidate responses $N$
    \end{itemize}
\Ensure Preference dataset $D$

\State $D \gets \emptyset$ \Comment{Initialize the preference dataset}
\State $C \gets \emptyset$ \Comment{Initialize the conversation history}
\State Initialize $C$ with the starting context

\While{length($C$) $<$ $L$}
    \State \textbf{Agent Decision Phase:}
    \State Generate $N$ candidate responses: $R \gets \{r_1, r_2, \dots, r_N\}$ from $A$
    \State $S \gets \emptyset$ \Comment{Initialize the list to store branch scores}
    
    \For{each response $r_i \in R$}
        \State \textbf{Initialize Branch:}
        \State $C_i \gets C$ \Comment{Clone the current conversation history}
        \State Append $r_i$ to $C_i$
        
        \State \textbf{Simulate Look-Ahead:}
        \State $steps \gets 0$
        \State $current\_turn \gets \text{User}$
        \While{$steps < K$ \textbf{and} termination condition not met}
            \If{$current\_turn = \text{User}$}
                \State $u \gets U(C_i)$ \Comment{Generate a user response}
                \State Append $u$ to $C_i$
                \State $current\_turn \gets \text{Agent}$
            \Else
                \State $a \gets A(C_i)$ \Comment{Generate an agent response}
                \State Append $a$ to $C_i$
                \State $current\_turn \gets \text{User}$
            \EndIf
            \State $steps \gets steps + 1$
        \EndWhile
        
        \State \textbf{Evaluate Branch:}
        \State Compute branch score $s_i \gets O(C_i)$
        \State Add $s_i$ to $S$
    \EndFor
    
    \State \textbf{Determine Preferences:}
    \State $w \gets \arg\max(S)$ \Comment{Index of the preferred response}
    \State $l \gets \arg\min(S)$ \Comment{Index of the least preferred response}
    \State Let $r_w \gets R[w]$ and $r_l \gets R[l]$
    
    \State \textbf{Record Preference Tuple:}
    \State $D \gets D \cup \{(C, r_w, r_l)\}$
    
    \State \textbf{Update Conversation History:}
    \State Append $r_w$ to $C$
    \State $u \gets U(C)$ \Comment{Generate the subsequent user reply}
    \State Append $u$ to $C$
    
    \If{termination condition is met}
        \State \textbf{Exit Loop:} \textbf{break}
    \EndIf
\EndWhile

\State \Return $D$
\end{algorithmic}
\end{algorithm}

\subsection{Evaluation Questionnaires for Therapist Performance}
\label{sec:Questionnaire}
\begin{table}[H]
    \centering
    \caption{The questions posed to the LLM for evaluating the performance of the therapist.}
    \begin{tabular}{p{0.05\textwidth}p{0.7\textwidth}}
        \toprule
        & \textbf{Questionnaire 1 (session satisfaction)}\\
        \midrule
        Q1 & Your overall satisfaction with the chat? \\
        Q2 & Your overall satisfaction with the content of the chat? \\
        Q3 & To what extent do you feel the chat facilitated motivation? \\
        Q4 & Did you learn anything? \\
        Q5 & To what extent was this learning relevant to your everyday life? \\
        \midrule
        & \textbf{Questionnaire 2 (working alliance)}\\
        \midrule
        Q1 & The therapist gave me a sense of who it was. \\
        Q2 & The therapist revealed what it was thinking. \\
        Q3 & The therapist shared its feelings with me. \\
        Q4 & The therapist seemed to know how I was feeling. \\
        Q5 & The therapist seemed to understand me. \\
        Q6 & The therapist put itself in my shoes. \\
        Q7 & The therapist seemed comfortable talking with me. \\
        Q8 & The therapist seemed relaxed and secure when talking with me. \\
        Q9 & The therapist took charge of the conversation. \\
        Q10 & The therapist let me know when it was happy or sad. \\
        Q11 & The therapist didn’t have difficulty finding words to express itself. \\
        Q12 & The therapist was able to express itself verbally. \\
        Q13 & I would describe the therapist as a “warm” communication partner. \\
        Q14 & The therapist did not judge me. \\
        Q15 & The therapist communicated with me as though we were equals. \\
        Q16 & The therapist made me feel like it cared about me. \\
        Q17 & The therapist made me feel close to it. \\
        \bottomrule
    \end{tabular}
    \label{Questionnaires}
\end{table}

\subsection{Tukey HSD Post-Hoc Analysis} 
\label{sec:tukey}
\begin{table}[H]
    \centering
    \caption{Tukey HSD Post-Hoc Test Results for Pairwise Model Comparisons}
    \label{tab:tukey}
    \begin{tabular}{lcccccc}
        \toprule
        \textbf{Comparison} & \textbf{Metric} & \textbf{Mean Diff} & \textbf{p-value} & \textbf{Lower Bound} & \textbf{Upper Bound} & \textbf{Sig.} \\
        \midrule
        Base vs. L0\_M4 & Final Score & 0.3235 & 0.0023 & 0.0987 & 0.5483 & Yes \\
        Base vs. L5\_M7 & Final Score & 0.5292 & $<$0.0001 & 0.3044 & 0.7540 & Yes \\
        L0\_M4 vs. L5\_M7 & Final Score & 0.2057 & 0.0807 & -0.0191 & 0.4305 & No \\
        \midrule
        Base vs. L0\_M4 & Q1 & 0.4479 & 0.0020 & 0.1407 & 0.7552 & Yes \\
        Base vs. L5\_M7 & Q1 & 0.6687 & $<$0.0001 & 0.3615 & 0.9760 & Yes \\
        L0\_M4 vs. L5\_M7 & Q1 & 0.2208 & 0.2095 & -0.0864 & 0.5281 & No \\
        \midrule
        Base vs. L0\_M4 & Q2 & 0.1991 & 0.0231 & 0.0221 & 0.3762 & Yes \\
        Base vs. L5\_M7 & Q2 & 0.3897 & $<$0.0001 & 0.2126 & 0.5668 & Yes \\
        L0\_M4 vs. L5\_M7 & Q2 & 0.1906 & 0.0315 & 0.0135 & 0.3676 & Yes \\
        \midrule
        Base vs. L0\_M4 & Conversation Length & -5.3438 & 0.0150 & -9.8378 & -0.8497 & Yes \\
        Base vs. L5\_M7 & Conversation Length & -9.2812 & $<$0.0001 & -13.7753 & -4.7872 & Yes \\
        L0\_M4 vs. L5\_M7 & Conversation Length & -3.9375 & 0.0992 & -8.4316 & 0.5566 & No \\
        \bottomrule
    \end{tabular}
\end{table}

\end{document}